%% file: acl_latex.tex
\pdfoutput=1

\documentclass[11pt]{article}

\usepackage[]{acl}

\usepackage{times}
\usepackage{latexsym}

\usepackage[T1]{fontenc}

\usepackage[utf8]{inputenc}

\usepackage{microtype}

\usepackage{inconsolata}

\usepackage{graphicx}

\usepackage{xcolor}  
\usepackage{pxrubrica}        
\usepackage{url}
\usepackage{booktabs}
\usepackage{multirow}
\usepackage{xspace}
\usepackage{amsmath}
\usepackage{subfigure}
\usepackage{comment}
\usepackage{makecell}
\usepackage{color}
\usepackage{pifont}
\usepackage{tabularx}
\usepackage{colortbl}
\usepackage{listings}
\usepackage{caption}
\usepackage{enumitem}
\usepackage{courier}

\usepackage{tikz}

\usepackage[whole]{bxcjkjatype} 
\usepackage{amssymb}
\usepackage{amsmath}
\usepackage{amsfonts}
\usepackage{latexsym}
\usepackage {tablefootnote} 
\definecolor{red}{rgb}{0.8,0,0}
\definecolor{green}{rgb}{0,0.8,0}
\definecolor{blue}{rgb}{0,0,0.8}
\definecolor{yellow}{rgb}{0.90,0.91,0.75}
\definecolor{purple}{rgb}{0.85,0.80,0.91}
\definecolor{navy}{rgb}{0.098,0.267,0.557}

\renewcommand{\paragraph}[1]{\noindent\textbf{#1} \hspace{2pt}}
\newcommand{\yes}{\ding{51}} %
\newcommand{\no}{\ding{55}} %
\newcommand{\breakInTable}[1]{\begin{tabular}{l}
    #1
\end{tabular}}

\newcommand{\navy}[1]{{\color{navy}#1}}
\title{Constructing Multimodal Datasets from Scratch \\ for Rapid Development of a Japanese Visual Language Model}

\author{
    Keito Sasagawa$^{\clubsuit,\ddag}$, 
    Koki Maeda$^{\diamondsuit,\ddag *}$,
    Issa Sugiura$^{\spadesuit,\ddag}$\thanks{Equally contributed.},\\
    \textbf{Shuhei Kurita}$^{\dag,\ddag}$,
    \textbf{Naoaki Okazaki}$^{\diamondsuit,\dag,\ddag}$,
    \textbf{Daisuke Kawahara}$^{\clubsuit,\dag,\ddag}$\\
  $^{\clubsuit}$Waseda University,
  $^{\diamondsuit}$Institute of Science Tokyo,
  $^{\spadesuit}$Kyoto University, \\
  $^{\dag}$National Institute of Informatics,
  $^{\ddag}$NII LLMC\\ }

\begin{document}
\maketitle
\begin{abstract}

To develop high-performing Visual Language Models (VLMs), it is essential to prepare multimodal resources, such as image-text pairs, interleaved data, and instruction data. While multimodal resources for English are abundant, there is a significant lack of corresponding resources for non-English languages, such as Japanese. To address this problem, we take Japanese as a non-English language and propose a method for rapidly creating Japanese multimodal datasets from scratch. We collect Japanese image-text pairs and interleaved data from web archives and generate Japanese instruction data directly from images using an existing VLM. Our experimental results show that a VLM trained on these native datasets outperforms those relying on machine-translated content.

\end{abstract}

\section{Introduction}

We develop a multimodal resource for high-performing Visual Language Models (VLMs) in Japanese.
While English multimodal resources are relatively abundant in existing studies, there is a significant lack of corresponding datasets for non-English languages, such as Japanese. One potential solution is translating multimodal resources from English into Japanese. However, this approach often produces suboptimal results, as the translation does not account for the contextual relationship between the image and the text. Such a translation approach also cannot follow the cultural backgrounds of the image domains as they are collected in English websites.

To address this multilingual gap, we propose a method for rapidly constructing Japanese multimodal datasets from scratch. We build two types of datasets: \textit{pretraining data} and \textit{instruction data} in Japanese. For the pretraining data, we curate a large-scale dataset of Japanese image-text pairs and interleaved data by extracting and localizing data from web crawls. For the instruction  tuning data, we follow the LLaVA~\cite{liu-etal-2023-LLaVA} methodology and directly generate Japanese instruction data by inputting Japanese images into an existing VLM via APIs, ensuring more accurate alignment between visual and textual content.

\begin{figure}[t]
    \centering
    \includegraphics[width=\linewidth]{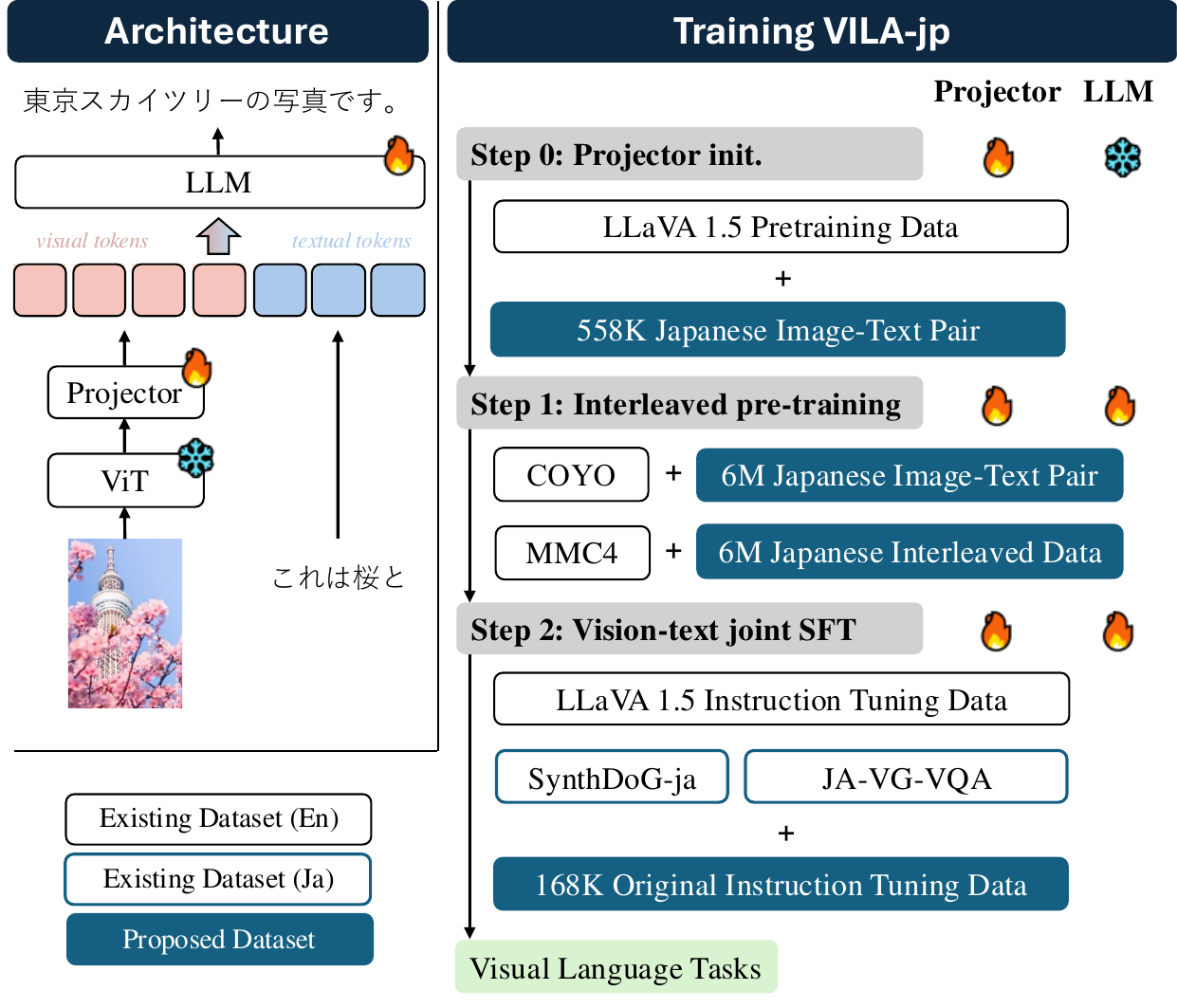}
    \caption{We propose \textbf{VILA-jp}, a novel Japanese vision and language model. For each step of pre-training and instruction tuning, we construct tailored million-scale image-text dataset (\navy{$\blacksquare$}) from interleaved data.}
    \label{fig:teaser}
    \vspace{-1em}
\end{figure}

We demonstrate that models trained on our proposed datasets achieve higher accuracy than those developed with machine-translated datasets, particularly for the instruction data. For this purpose, we adopt the VILA~\cite{vila} architecture for our Japanese VLM with English publicly available data and Japanese newly collected data.
We assume that this approach is adaptable to any languages and not limited to Japanese. Our contributions significantly enhance the resources available for Japanese VLM studies, enabling more effective regional localization and cultural understanding in VLMs.

\section{Related Work}
\paragraph{Resource for Visual Language Models}
Visual language models require a large amount of image-text pairs for both pretraining and instruction tuning.
Visual instruction tuning of LLaVA~\cite{liu-etal-2023-LLaVA,liu-etal-2023-LLaVA1.5} relies on synthesized data by OpenAI GPT~\cite{gpt3}.
This synthesized instruction data is used in the various VLM developments such as InstructBLIP~\cite{instructblip} and VILA~\cite{vila}. Following these successes, we examine the use of the synthesized data in instruction tuning in Japanese.

\paragraph{Japanese Visual Language Resources}
The task of answering questions about documents that include both visual and textual contents has garnered attention as Document Visual Question Answering (Document VQA)~\cite{AI2D2016,TQA2017,FOODWEDS2016, DVQA2018, VLQA2020, SlideVQA2023}.
However, these datasets are mostly developed in English and do not reflect the domain-specific knowledge in other languages.
Recently, JDocQA was proposed for a Japanese document visual question answering~\cite{jdocqa}.
Heron Bench~\cite{inoue-etal-2024-heronbench} was also proposed for evaluating Japanese VLM ability.
However, Japanese resources are quite limited compared to English resources, and existing ones are often intended for fine-tuning models.
In this paper, we construct a large-scale text-image corpora for Japanese VLM construction.

\section{Dataset Construction}
\label{sec:dataset_construction}

\begingroup
\renewcommand{\arraystretch}{0.9}
\begin{table}[t]
\centering
\resizebox{\columnwidth}{!}{%
\begin{tabular}{lccc}
\toprule
\textbf{Data} & \textbf{\# Images} & \textbf{\# Step} & \textbf{Full} \\
\midrule
\textit{English} \\
LLaVA-1.5 pretrain data          & 558K & 0   & \yes \\
LLaVA-1.5 instruction data (subset)   & 358K & 2   & \yes \\
COYO~\cite{kakaobrain2022coyo-700m} (subset)          & 6M & 1 & \yes \\
mmc4-core~\cite{multimodalc4} (subset)     & 6M & 1 & \yes \\
\midrule
\textit{Japanese} \\
Translated data\tablefootnote{https://huggingface.co/datasets/turing-motors/LLaVA-v1.5-Instruct-620K-JA}  & 620K   & 2 & \no \\
(Proposed) Image-text pairs   & 6.6M   & 0 \& 1 & \yes \\
(Proposed) Interleaved        & 6M & 1 & \yes\\
(Proposed) Instruction tuning & 369K   & 2 & \yes \\
\bottomrule
\end{tabular}
}
\caption{Data size for VILA-jp. Full means the dataset is used in our full model.}
\label{table:dataset_split}
\end{table}
\endgroup

\paragraph{Interleaved data}
This dataset is constructed based on the data construction process of MMC4~\cite{multimodalc4}.
The texts in the dataset come from Japanese texts extracted from the 2020-2022 Common Crawl dumps in the llm-jp-corpus~\cite{llmjp}.
We use \texttt{bunkai}~\cite{hayashibe-mitsuzawa-2020-sentence} to break down Japanese text into sentences.
After this process, sentences consisting only of symbols with no numbers, English or Japanese characters were combined with the previous sentence.
In addition, if the end of the parentheses come at the beginning of the next sentence, it is moved to the end of the previous sentence.

We download images from URLs that are extracted from web texts.
To avoid overloading on specific servers, we downsample URLs from frequent domains.
Next, we remove duplicate images within the same document.
We use \texttt{ImageHash}\footnote{https://github.com/JohannesBuchner/imagehash} to calculate the phash value of the image, and for images with a Hamming distance of 5 or less, we keep the one with the highest resolution.
We also remove duplicate images across multiple documents. For data from each year, we remove images that have more than 10 duplicates in the 60K images sampled. This operation is repeated until the total number of sampled images is the same as the original number of images. This removes application icons, advertisements, and images that are inserted when links are broken.
Then, we use the NSFW image classifier from \texttt{dataset2metadata}\footnote{https://github.com/mlfoundations/dataset2metadata} to remove NSFW images~\cite{NEURIPS2023_56332d41}.
We delete images with a classifier output score of 0.1 or higher.

\input{tables/table_main}

For images that have passed through these filters, we calculate the similarity of all pairs of images and sentences in the document using the OpenCLIP~\cite{ilharco_gabriel_2021_5143773} trained on LAION5B dataset\footnote{https://huggingface.co/laion/CLIP-ViT-H-14-frozen-xlm-roberta-large-laion5B-s13B-b90k}~\cite{NEURIPS2022_a1859deb}.
We remove images that do not have a CLIP similarity of at least 0.20 with any of the sentences in the document.
In the same way as the construction method for MMC4, we map images to text by solving an assignment problem using \texttt{lapjv}\footnote{https://pypi.org/project/lapjv/}.
Finally, we use the harmful text filtering applied in llm-jp-corpus.
Furthermore, we only kept samples with the number of images between 2 and 5, the number of sentences between 10 and 100, the token length of the sample within the max length of the LLM, and the similarity of all assigned image and text pairs above 0.20.
As a result, the number of images in the dataset is 9.9M. For training, we use a subset of this dataset to balance the Japanese image-text pair dataset.

\paragraph{Image-text pairs}
We collected alt attributes for images after NSFW image filtering in interleaved data.
We performed text filtering, based on the filtering method used in constructing the COYO dataset~\cite{kakaobrain2022coyo-700m}.
First, we use a regular expression to remove all text that does not contain any hiragana, katakana, or common kanji characters.
We also filter too short alt texts and specific file names of images that are typically for ads or screenshots.
Next, we filter NSFW content using the DiscardAdultContentJa filter of the \texttt{Hojichar}\footnote{github.com/HojiChar/HojiChar}.
For text that pass through these filters, the first and last consecutive whitespace characters are removed, and if there are two or more consecutive whitespace characters, they are replaced with a half-width space.
Then we deduplicate the data.
Alt text that appeared more than 10 times was removed, and duplicates were removed for (image phash value, alt text) pairs.

Finally, the similarity of each image and alt text pair is calculated using OpenCLIP trained on LAION5B dataset and Japanese CLIP\footnote{huggingface.co/line-corporation/clip-japanese-base}.
We also filter lower 30 percentile of CLIP alignment score data.
The resulting dataset contains 6.6M images.

\begin{figure*}[h]
    \centering
    \includegraphics[width=.93\linewidth]{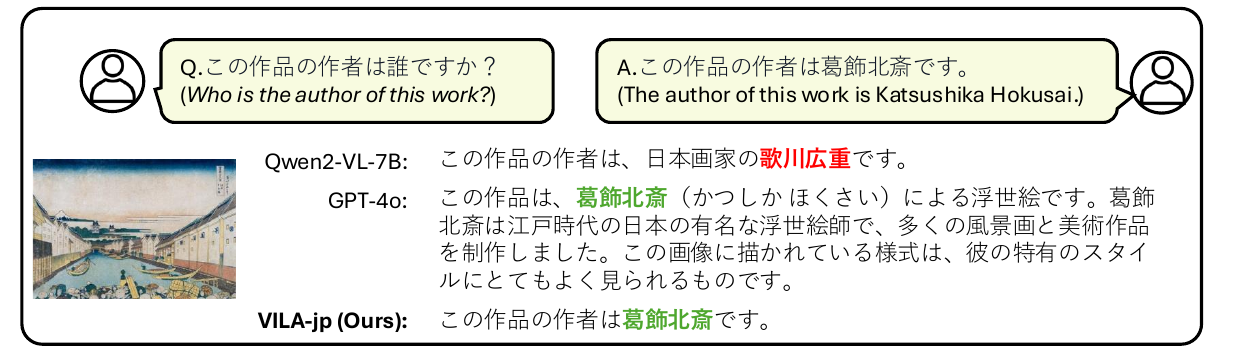}
    \caption{Examples of text generated by each model in response to a question from the Japanese-Heron-Bench. \textcolor{green}{\textbf{Green}} indicates the correct word and \textcolor{red}{\textbf{red}} indicates the wrong word.}
    \label{fig:example-heron}
\end{figure*}

\paragraph{Instruction Data}
There is already a Japanese multimodal instruction dataset, the LLaVA-Instruct dataset translated into Japanese using DeepL, but it contains unnatural Japanese due to translation errors.
Using Azure OpenAI API, we construct a Japanese dataset from COCO images~\cite{coco} in the same way as the LLaVA-Instruct dataset.
As a result, we have a dataset of 156K samples, including conversation, detail, and complex reasoning types.

In addition, we use GPT-4o to generate multi-turn conversation data from the Japan Diverse Images Dataset\footnote{huggingface.co/datasets/ThePioneer/japanese-photos} and develop \textbf{Japanese-photos-conv} dataset. This dataset has 12K samples. We summarize the dataset used in VILA-jp in Table~\ref{table:dataset_split} and details in Appendix~\ref{appendix:japanese_photos_conv}.

\input{tables/table_ablation}

\section{Experiments}
\label{sec:experiment}

\subsection{Model Training}
Our model architecture integrates the vision encoder and LLM through the projector, similarly to VILA.
We use SigLIP\footnote{huggingface.co/google/siglip-so400m-patch14-384}~\cite{zhai2023sigmoid} for vision encoder, llm-jp-3-13b-instruct\footnote{huggingface.co/llm-jp/llm-jp-3-13b-instruct} for LLM, and two-layer MLP for projector.

We train our model in three training stages, inspired by VILA as shown in Figure~\ref{fig:teaser}.
The first stage is the projector initialization stage, where only the projector parameters are tuned on English and Japanese image-text pair datasets.
We use 558K samples of English and Japanese image-text pairs.
The second stage is a multimodal continual pretraining stage, in which the parameters of the projector and LLM are tuned on image-text pair datasets and interleaved datasets.
For the English dataset, we use a subset of 6M images from mmc4-core and a subset of 6M images from coyo. For the Japanese datasets, we use our pair dataset and our interleaved dataset, each with 6M images.
The third stage is the multimodal instruction tuning stage, where the projector and LLM are tuned so that the model can follow instructions. We used JA-VG-VQA~\cite{shimizu-etal-2018-japanesevgvqa} and synthdog-ja~\cite{kim2022ocr} datasets in addition to the datasets in Sec.~\ref{sec:dataset_construction} as Japanese datasets, and a subset of LLaVA-Instruct as English dataset. The Japanese instruction dataset has a total of 369K samples, and the English instruction dataset has a total of 358K samples.
Details of the datasets are described in the Appendix~\ref{appendix:training_dataset}.

\subsection{Evaluation}
To verify the comprehensive ability of VILA-jp, we employed three benchmarks: Heron Bench~\cite{inoue-etal-2024-heronbench}, JA-VLM-Bench-In-the-Wild~\cite{akiba-etal-2024-evomodelmerge}, and JA-VG-VQA500\footnote{See Appendix~\ref{app:bench-details} for benchmark details and Appendix~\ref{app:eval-details} for evaluation methodologies.}.
In this evaluation, we used the LLM-as-a-Judge framework~\cite{zheng2023judging}. 

\paragraph{Benchmark Results}
Table~\ref{table:main} presents the performance on three benchmarks. 
Compared to current VLMs of similar size, VILA-jp consistently achieved state-of-the-art performance, as measured by the LLM-as-a-Judge score. 
Additionally, on the JA-VG-VQA-500 benchmark, our model surpassed even the performance of GPT-4o.
While ROUGE-L is commonly used for evaluation, it tends to deteriorate when the generated output deviates significantly from the provided reference. This is evidenced by GPT-4o's low ROUGE-L score despite its demonstrably high capabilities, suggesting that ROUGE-L may no longer be a suitable metric for evaluating performance in this context.

\paragraph{Qualitative Evaluation}
Figure \ref{fig:example-heron} shows the qualitative comparisons between VILA-jp and existing models. While Qwen2-VL-7B misidentifies the answer, confusing it with another famous Japanese artist 歌川広重 \textit{(Utagawa Hiroshige)}, our model precisely answers the question, showing capability for Japanese culture-specific knowledge. GPT-4o also provides a good answer with rich information. 
We provide additional examples in Appendix~\ref{sec:app-qual-example}.

\paragraph{Ablation of the training dataset}
To validate the effectiveness of our constructed dataset, we did ablation study of the constructed dataset.
We ablated our instruction data, step 1 training, and our interleaved data.
Table~\ref{table:ablation} highlights that the use of translated instruction data degrades performance significantly compared to our proposed dataset.
While there is a some improvement with step 1 training, the effect of interleaved data is still unclear.

\section{Conclusion}
We introduced VILA-jp, a Japanese Visual Language Model designed to integrate multiple images with natural language understanding. Our experiments demonstrate the model's effectiveness across a range of multimodal tasks in Japanese. For future work, we plan to extend our model to wide branch of Japanese visual datasets.

\section*{Acknowledgments}
This research project has benefited from the Microsoft Accelerate Foundation Models Research (AFMR) grant program for the use of GPT-4o through which leading foundation models hosted by Microsoft Azure along with access to Azure credits were provided to conduct the research.



\section*{Limitation}
Since we generate training dataset with proprietary models, the quality relies on such models.
We need to find ways to synthesize better quality, less hallucinated datasets.
Also, visual knowledge of Japan may depend on the original vision encoder.
We may need to build another vision encoder from scratch or conduct additional tuning.

\section*{Ethical Consideration}
We are implementing NSFW filtering for images and text, so it is thought that there are almost no such images.
In addition, when generating data using GPT-4o, NSFW images are rejected by Azure OpenAI, so it is likely that there will be almost no such images in the generated instruction data.

\bibliography{custom}

\clearpage
\appendix

\section{Dataset Construction Details}
\subsection{Image Downloading}

When we download images from URLs, we limit the extensions of the image file to ``.jpg'', ``.jpeg'', and ``.png'', and if the words ``logo'', ``button'', ``icon'', ``plugin'', or ``widget'' is included, we skip them.
If the size of the image is less than 150px in either the height or width, it will be removed. 
Also, if the aspect ratio of the image is less than 0.5 or greater than 2, it will be removed.

\subsection{Filtering for Image-text Pair Dataset}

Some data has text that is set when the alt attribute is not set.
We remove such data, where the text begin with ``画像に alt 属性が指定されていません。'' or ``この画像には alt 属性が指定されておらず、''.

In addition, some data have the image file name set automatically when a screenshot is taken, etc., as alt text.
For example, ``写真 2015-01-20 18 12 33''.
Specifically, if the data does not contain Japanese after ``写真,'' ``キャプチャ,'' ``画像,'' ``スクリーンショット,'' ``全画面キャプチャ,'' ``ファイル,'' ``コメント,'' or ``コピー,'' the data will be removed.

\subsection{Japanese Photos Conv Dataset}
\label{appendix:japanese_photos_conv}
Japan Diverse Images Dataset\footnote{huggingface.co/datasets/ThePioneer/japanese-photos} consists of images taken in Japan.
For each image in this dataset, we generate a multi-turn question answer via \texttt{gpt-4o-2024-05-13}.
In generating, we adopted zero-shot manner with the image as input. The system prompt for QA generation is shown in Table~\ref{tab:system-prompt-for-japanese-photos}.
Except several images filtered by Azure OpenAI, we collected dataset with 12K samples.

\begin{table*}[h]
    \centering
    \footnotesize
    \caption{
      The prompt used in generating japanese-photos-conv.
    }
    \label{tab:system-prompt-for-japanese-photos}
    \scalebox{1}{
        \begin{tabularx}{\linewidth}{X}
          \toprule
            \textbf{Prompt} \\
          \midrule
            \input{tables/prompt_japanese_photos} \\
          \bottomrule
        \end{tabularx}
    }
\end{table*}

\section{Existing Training Dataset Details}
\label{appendix:training_dataset}
\paragraph{JA-VG-VQA}

For training, we used 99K samples, excluding 500 samples from JA-VG-VQA-500.
Since multiple QAs are assigned to each image, they are combined to create multi-turn conversation data.
Also, since the answers in this dataset are phrases or short sentences, we specify the response format to be that way. Specifically, we add ``語句または短い文で答えてください。(\textit{Please answer in phrases or short sentence.})'' to the first question.

\paragraph{SynthDoG-ja}

We use a subset of 102K samples to match the data volume of the English OCR task dataset.
We prepare several templates and converte the dataset into a QA format.

\paragraph{English instruction dataset}

For the English dataset, a subset of the instruction data from LLaVA-1.5 is used for training to match the amount of data in the Japanese dataset.
We use 158K samples from the LLaVA-Instruct dataset as a synthetic dataset with GPT-4.
As VQA datasets, we use a 53K sample subset of VQAv2~\cite{goyal2017making} and a 46K sample subset of GQA~\cite{hudson2019gqa}, which are multi-turn datasets similar to ja-vg-vqa.
We also use the OCRVQA~\cite{mishra2019ocr} dataset with 80K samples and the TextCaps~\cite{sidorov2020textcaps} dataset with 22K samples as datasets for the English OCR task.
In total, these datasets result in 358K samples of English instruction data.

\section{Training Details}
\paragraph{Model Parameters}
The number of model parameters for each module is shown in the table~\ref{table:model_parameters}.
\begin{table}[t]
\centering
\begin{tabular}{lc}
\toprule
\textbf{Module} & \textbf{\# Params} \\
\midrule
Vision Encoder   & 428M  \\
Projector   & 32M  \\
LLM          & 13B  \\
\bottomrule
\end{tabular}
\caption{Model parameters for VILA-jp.}
\label{table:model_parameters}
\end{table}

\paragraph{Computational Budget}
Training for step 0 takes about 14-15 hours on 1 node with 8xA100 (40GB).
Step 1 takes about 130 hours to training on 8 nodes with 8xA100 (40GB).
Step 2 takes about 11 hours to training on 4 nodes with 8xA100 (40GB).

\paragraph{Hyperparameters}
Table~\ref{table:training_hyperparameters} shows the hyperparameters for each training step.

\begin{table}[t]
\centering
\begin{tabular}{l|ccc}
\toprule
\textbf{Hyperparameter} & \textbf{Step 0} & \textbf{Step 1} & \textbf{Step 2}\\
\midrule
batch size   & 256 & 1024 & 128 \\
larning rate (lr)   & 1e-3 & 5e-5 & 1e-5 \\
lr scheduler type  & \multicolumn{3}{c}{cosine} \\
lr warmup ratio    & \multicolumn{3}{c}{0.03}  \\
weight decay    & \multicolumn{3}{c}{0} \\
epoch    &  \multicolumn{3}{c}{1} \\
optimizer    &   \multicolumn{3}{c}{AdamW}  \\
DeepSpeed Stage   &  2    & \multicolumn{2}{c}{3}  \\
\bottomrule
\end{tabular}
\caption{Hyperparameters for each training step}
\label{table:training_hyperparameters}
\end{table}

\section{Baseline Models}
\label{sec:app-baseline}

We present a overview of the vision language models and their corresponding base language models, sizes, and Hugging Face repositories or APIs in Table~\ref{table:model_information}.

\section{Benchmark Dataset Details}
\label{app:bench-details}

Here we provide detail information of the dataset used in the experiments (Sec.~\ref{sec:experiment}).

\paragraph{Heron Bench}
evaluates the Japanese language capabilities of VLMs using a dataset of 21 images and 103 image-question-answer triplets designed specifically within cultural and linguistic contexts of Japan.

\paragraph{JA-VLM-Bench-In-the-Wild} comprises 42 images paired with 50 curated questions focusing on a diverse range of culturally specific elements and objects commonly found in Japan. In developing the benchmark, authors leveraged GPT-4V~\cite{openai-2024-gpt4} and conducted human-in-the-loop filtering process to ensure the dataset quality.

\paragraph{JA-VG-VQA500\footnote{huggingface.co/datasets/SakanaAI/JA-VG-VQA-500}} is a subset of the Japanese Visual Genome VQA dataset~\cite{shimizu-etal-2018-japanesevgvqa}, which is based on Visual Genome~\cite{krishna-etal-2016-visualgenome}, extracted 500 samples from the test set.

\section{Evaluation Details}
\label{app:eval-details}

We largely follow the original evaluation settings; however, we have introduced certain modifications to the evaluation methods to better reflect the objectives of our study.
We averaged scores of 5runs in Heron Bench and JA-VLM-Bench-In-the-Wild and employed a single run score of JA-VG-VQA-500.

\paragraph{Heron Bench:}
We followed LLM-as-a-judge approach via Azure OpenAI API and employed \texttt{gpt-4o-2024-05-13}.
The model's performance is quantified by the ratio of its average score of answers evaluated in LLM-as-a-judge process to that obtained by GPT-4.
Consequently, scores can exceed 100\%, which indicates that the model outperformed GPT-4 on average.
For reproducibility, the evaluator's temperature was set to 0 and the seed to 0. This setting is also the case for other LLM-as-a-Judge-based evaluations. 
Note that even the same seed is used, the output may not be deterministic\footnote{For more information, refer to \url{https://platform.openai.com/docs/advanced-usage/reproducible-outputs}}.

\begin{figure}[t]
    \centering
    \includegraphics[width=\linewidth]{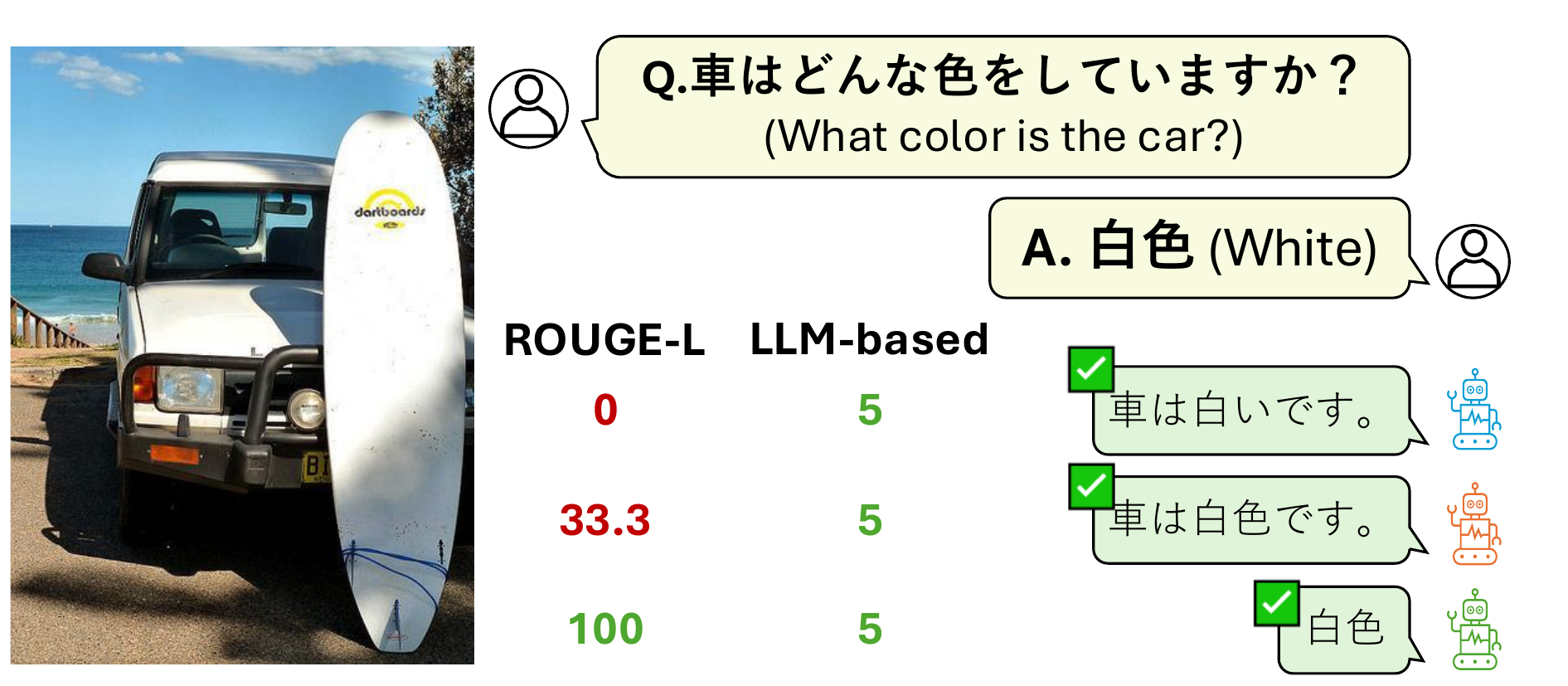}
    \caption{Example of how ROUGE-L scores can be misleading when evaluating Japanese VQA responses. Three different LLM-generated answers are shown alongside their corresponding ROUGE-L scores. }
    \label{fig:bad-rouge-example}
\end{figure}

\paragraph{Metrics:}
In \textbf{JA-VLM-Bench-In-the-Wild} and \textbf{JA-VG-VQA500}, ROUGE-L~\cite{lin-2004-rouge} is commonly used for the evaluation metrics.
However, the value of ROUGE varies greatly depending on the style of the answer in Japanese question answering. 
Figure~\ref{fig:bad-rouge-example} presents a typical example where ROUGE scores vary significantly due to differences in sentence structure and wording. VLMs are asked to identify the color of a car in an image, and the reference answer is ``白色 (White)''. Three different answers are shown, each with varying levels of detail and grammatical structure:
\begin{enumerate}
\setlength{\itemsep}{-4pt}
    \item ``車は白い です。'' (The car is white.)
    \item ``車は白色です。'' (The car is white.)
    \item ``白色'' (White)
\end{enumerate}
While all three answers are factually correct, their ROUGE-L scores differ significantly. The simplest answer received the highest ROUGE-L score, even though it lacks the grammatical completeness. In contrast, other answers got a score of 0, despite conveying the same information.
This example highlights the limitations of ROUGE-L in capturing the semantic nuances and stylistic variations of Japanese language. It suggests that relying solely on ROUGE-L might lead to an inaccurate assessment of LLM performance in VQA tasks, particularly in languages where word order is less rigid and contextual understanding is crucial.

To prevent such underestimation, we performed the LLM-as-a-judge process with the help of GPT-4o.
we evaluate the generated response in five-point Likert scale with some modification to a standard VQA prompt template, which is publicly available\footnote{\url{https://cloud.google.com/vertex-ai/generative-ai/docs/models/metrics-templates\#pointwise_question_answering_quality}}.
We present the actual prompt used in the evaluation in Table~\ref{tab:app-prompt-llm-as-a-judge}.

\paragraph{Excluding JMMMU}
We are aware of the JMMMU~\cite{onohara-etal-2024-jmmmu}, a valuable resource for evaluating Japanese vision and language models. However, our instruction tuning process for VILA-jp focused on generating free-form answers rather than selecting from a predefined set of options. As JMMMU primarily consists of multiple-choice questions, it was deemed unsuitable for assessing the performance of our model in its current training stage. We plan to explore fine-tuning strategies specifically for multiple-choice QA in future work.

\section{Detailed Result of Heron Bench}
\label{sec:app-heron}

Table~\ref{table:heron} provides a breakdown of Heron-Bench scores by category for each model. Our model demonstrates strong performance across all categories.

\input{tables/model_infomation}

\input{tables/prompt_llm_as_a_judge}

\input{tables/table_heron}

\section{Additional Qualitative Examples}
\label{sec:app-qual-example}

\begin{figure*}[t]
    \centering
    \includegraphics[width=\linewidth]{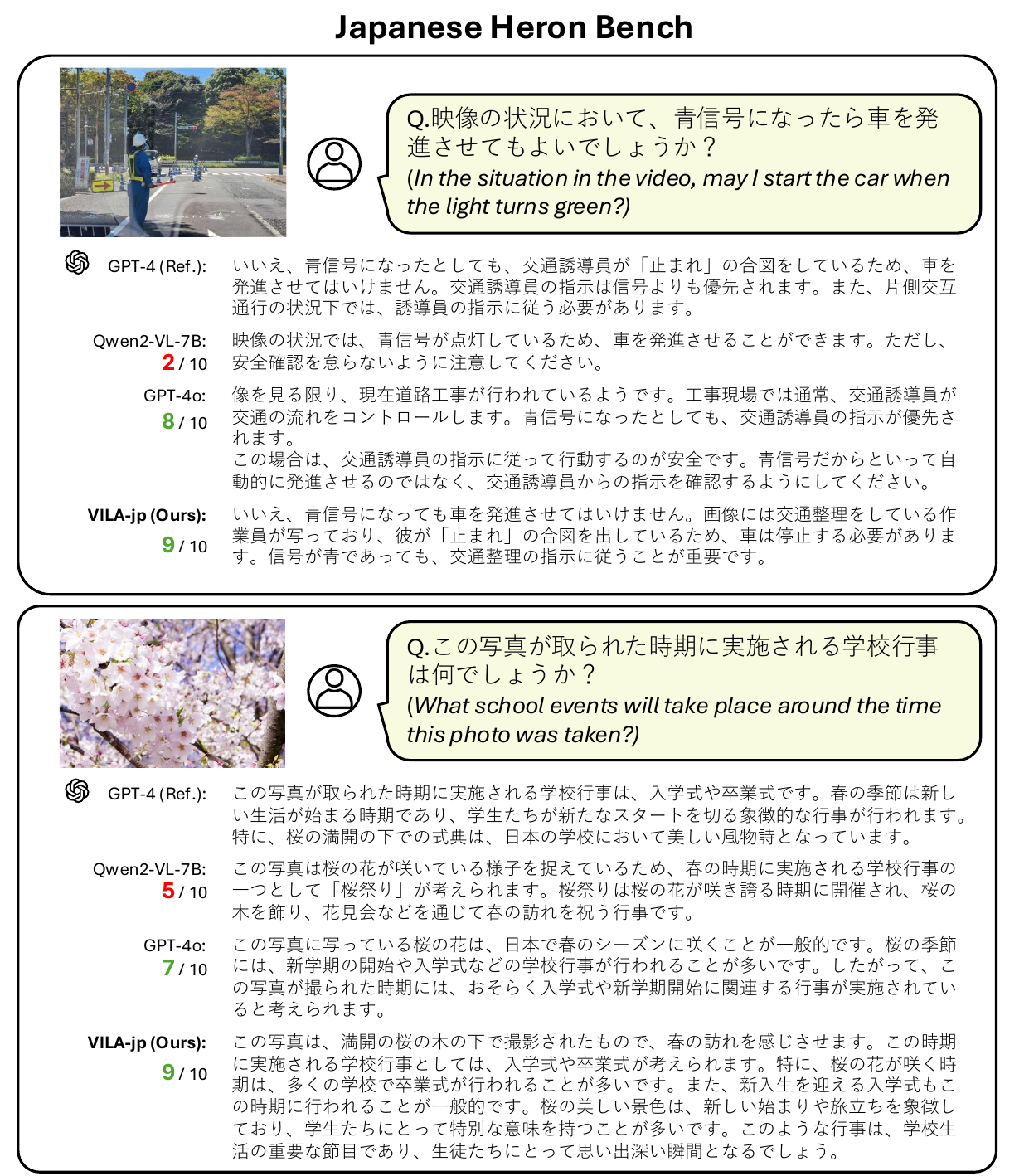}
    \caption{Additional qualitative examples in the Heron Bench.}
    \label{fig:add-example-heron-1}
\end{figure*}

\begin{figure*}[t]
    \centering
    \includegraphics[width=\linewidth]{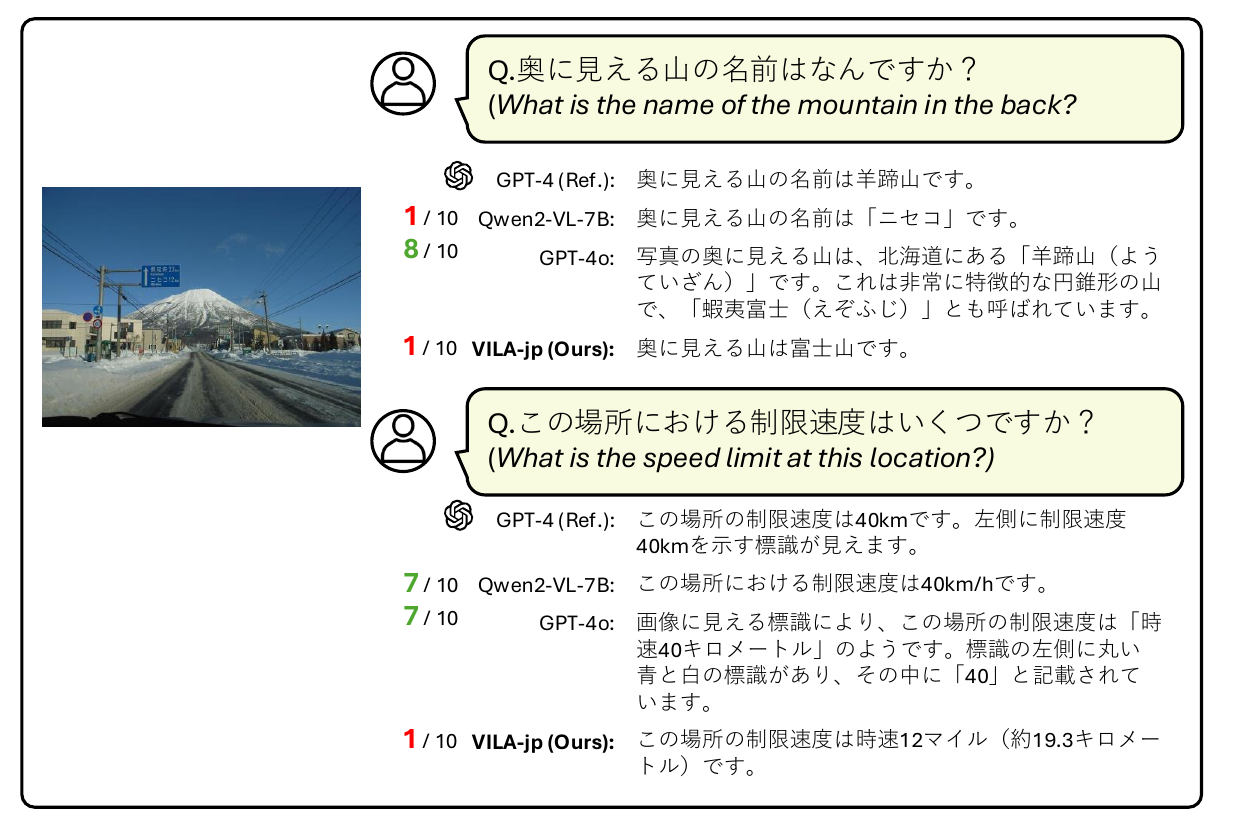}
    \caption{Continuous qualitative examples in the Heron Bench.}
    \label{fig:add-example-heron-2}
\end{figure*}

\begin{figure*}[t]
    \centering
    \includegraphics[width=\linewidth]{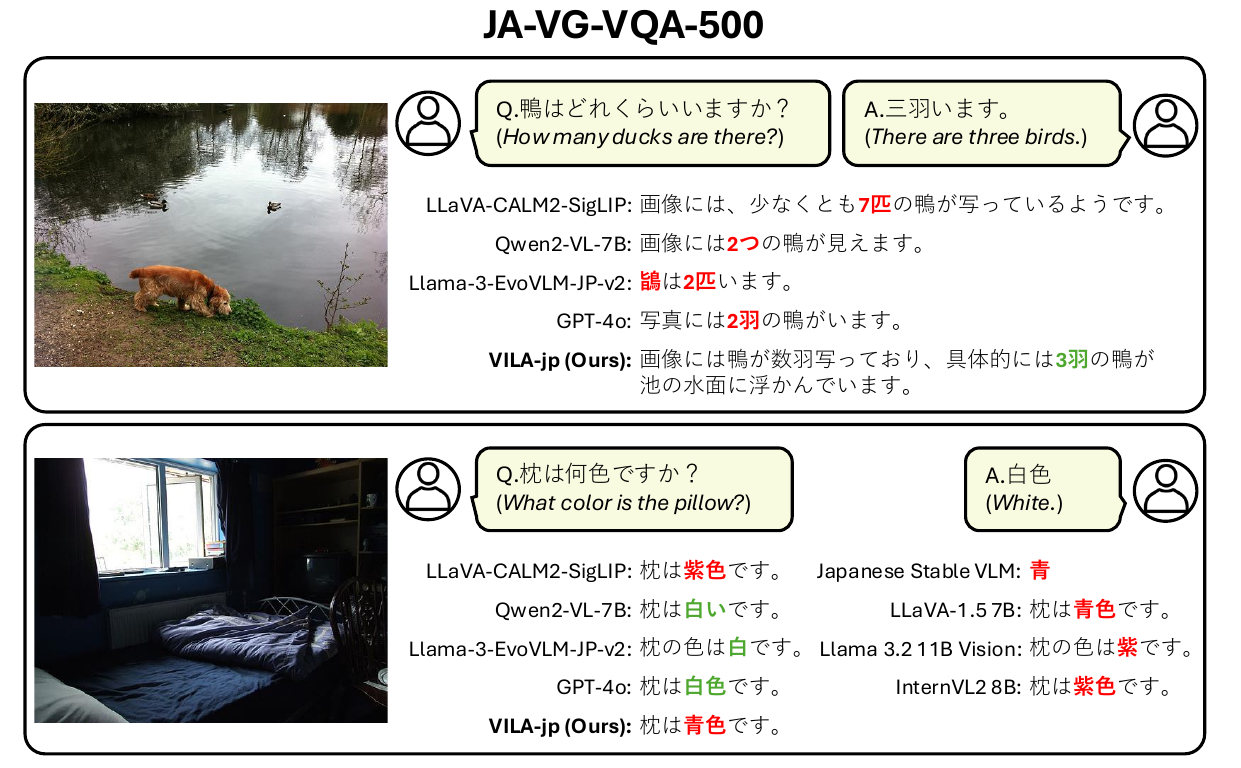}
    \caption{Additional qualitative examples in the JA-VG-VQA-500.}
    \label{fig:add-example-vqa}
\end{figure*}

\begin{figure*}[t]
    \centering
    \includegraphics[width=\linewidth]{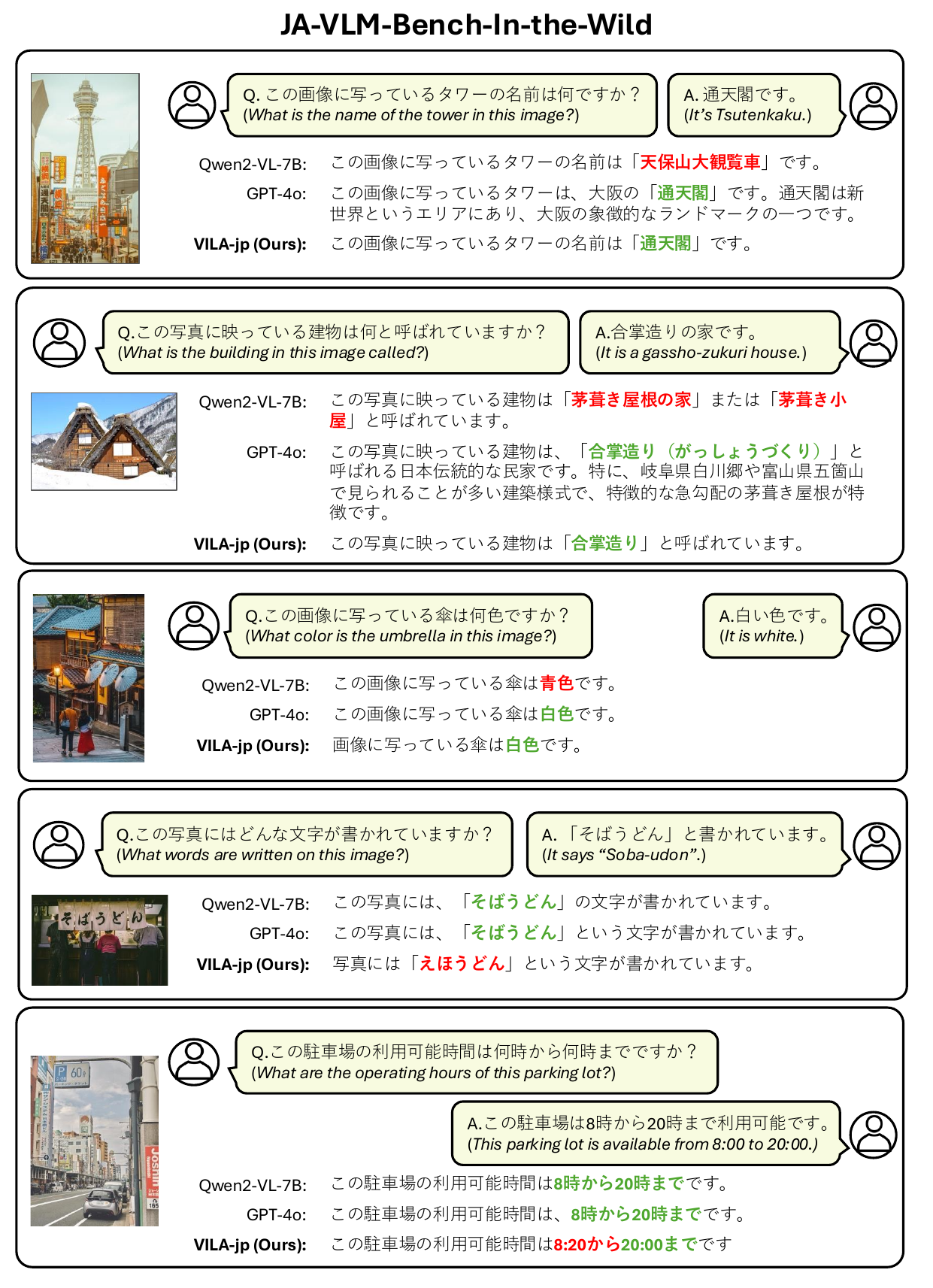}
    \caption{Additional qualitative examples in the JA-VLM-Bench-In-the-Wild.}
    \label{fig:add-example-in-the-wild}
\end{figure*}

Here we provide additional examples in Figure~\ref{fig:add-example-heron-1},\ref{fig:add-example-heron-2},\ref{fig:add-example-vqa}, and \ref{fig:add-example-in-the-wild}.

\end{document}

%% file: tables/table_main.tex
\begingroup
\renewcommand{\arraystretch}{0.9}
\begin{table*}[t]
\centering
\small
\begin{tabular}{lccccc}
\toprule
& \textbf{Heron-Bench} & \multicolumn{2}{c}{\textbf{JA-VLM-Bench-In-the-Wild}} & \multicolumn{2}{c}{\textbf{JA-VG-VQA-500}}  \\
\cmidrule(lr){2-2} \cmidrule(lr){3-4} \cmidrule(lr){5-6}
\textbf{Models} & LLM (\%) & ROUGE-L & LLM (/5.0) & ROUGE-L & LLM (/5.0) \\
\midrule
Japanese InstructBLIP Alpha & 14.0 & 20.8 & 2.42 & -- & -- \\ 
Japanese Stable VLM         & 24.2 & 23.3 & 2.47 & -- & -- \\ 
Llama-3-EvoVLM-JP-v2        & 39.3 & 41.4 & 2.92 & \textbf{23.5} & 2.96 \\
LLaVA-CALM2-SigLIP          & 43.3 & 47.2 & 3.15 & 17.4 & 3.21 \\
\midrule
LLaVA-1.6 7B                & 25.8 & 28.6 & 2.40 & 11.7 & 2.67 \\
LLaVA-1.5 7B                & 34.8 & 40.6 & 2.48 & 13.9 & 2.66 \\
Llama 3.2 11B Vision        & 36.5 & 27.4 & 2.77 & 13.8 & 2.95 \\
InternVL2 8B                & 45.2 & 33.7 & 2.98 & 11.6 & 3.13 \\
Qwen2-VL 7B Instruct        & 54.8 & 45.3 & 3.53 & 16.2 & 3.48 \\
\midrule
\textbf{VILA-jp (Ours)}      & \textbf{57.2} & \textbf{52.3} & \textbf{3.69} & 16.2 & \textbf{3.62} \\
\midrule
GPT-4o                      & 87.6 & 37.6 & 3.85 & 12.1 & 3.58 \\
\bottomrule
\end{tabular}
\caption{Comparison on Japanese benchmarks between current VLMs and \textbf{VILA-jp}. "--" indicates that the score cannot be calculated as the benchmark dataset is used for training. \textbf{Bold} indicates the best score except GPT-4o. "LLM" is an abbreviation for LLM-as-a-Judge. Detail information baseline models are in Appendix~\ref{sec:app-baseline} and a breakdown of Heron Bench scores by category in Appendix~\ref{sec:app-heron}.
}
\label{table:main}
\vspace{-1em}
\end{table*}
\endgroup

%% file: tables/table_ablation.tex
\begingroup
\renewcommand{\arraystretch}{0.9}
\begin{table*}[t]
\centering
\small
\begin{tabular}{cccccccc}
\toprule
 &&& \textbf{Heron-Bench}  & \multicolumn{2}{c}{\textbf{JA-VLM-Bench-In-the-Wild}} & \multicolumn{2}{c}
 {\textbf{JA-VG-VQA-500}}  \\
 \cmidrule(lr){4-4} \cmidrule(lr){5-6} \cmidrule(lr){7-8}
Step-0 & Step-1 & Step2 & LLM (\%) & ROUGE-L & LLM (/5.0) & ROUGE-L & LLM (/5.0) \\
\midrule
 \yes & \yes &translated   & 47.2 & 45.6 & 3.19 & 15.7 & 3.33 \\
 \yes & \no & \yes & 56.5 & \textbf{57.3} & 3.47 & 16.1 & 3.54 \\
 \yes & w/o interleaved & \yes & \textbf{58.6} & 52.2 & 3.50 & \textbf{16.7} & 3.61 \\
 \yes & \yes & \yes  & 57.2 & 52.3 & \textbf{3.69} & 16.2 & \textbf{3.62} \\
\bottomrule
\end{tabular}
\caption{\textbf{Ablation study of VILA-jp.} \textbf{Bold} indicates the best score. ``LLM'' is an abbreviation for LLM-as-a-Judge.}
\label{table:ablation}
\vspace{-1em}
\end{table*}
\endgroup

%% file: tables/prompt_japanese_photos.tex
\begin{verbatim}
あなたは、マルチモーダルinstruction tuningデータの優秀なアノテーターです。

これから、1枚の画像が与えられるので、その画像に関するinstruction tuningのための高品質なデータセットを生成する必要があります。このデータは、モデルが異なるモーダリティ間での関連性や相互作用を学習できるよう設計されなければなりません。

データにはオブジェクトの種類、オブジェクトの数、オブジェクトの動作、オブジェクトの位置、オブジェクト間の相対位置など、画像の内容を尋ねる質問を含めてください。また、明確な答えがある質問のみを含め、自信を持って答えられない質問はしないようにしてください。

また、画像の内容に関連した複雑な質問、例えば、画像に写っているオブジェクトの背景知識を尋ねる質問、画像の中で起こっている出来事について議論するよう求める質問なども含めてください。この場合も、不確かな詳細については質問しないようにしてください。
複雑な質問に回答する際は、詳細な回答にしてください。例えば、詳細な例や推論の手順を示すことで、内容に説得力を持たせ、よく整理された回答にすることができます。

出力形式は次のようにしてください。
```
Q:
{質問}
A:
{回答}
===
Q:
{質問}
A:
{回答}
===
・・・(省略)
===
Q:
{質問}
A:
{回答}
```
\end{verbatim}

%% file: tables/model_infomation.tex
\begin{table*}[t!]
\centering
\small
\caption{Comparison of Vision Language Models (VLMs) highlighting their base language models and its sizes (in billions of parameters), and corresponding Hugging Face repositories or API.}
\scalebox{0.8}{
    \begin{tabular}{lllrl}
    \toprule
    \textbf{VLM} & \textbf{Reference} & \textbf{Base LM} & \textbf{LM Size} & \textbf{Hugging Face / API}\\
    \midrule
     Llama 3.2 Vision 11B & \cite{meta-2024-llama-3.2-11B-vision} & Llama 3.2 & 11B & meta-llama/Llama-3.2-11B-Vision \\
     Qwen2-VL 7B & \cite{wang-etal-2024-qwen2vl}& Qwen2 & 7B & Qwen/Qwen2-VL-7B-Instruct\\
     InternVL2 8B & \cite{chen-etal-2024-internvl1.5} & InternLM2-Chat & 8B & OpenGVLab/InternVL2-8B \\
     LLaVA-1.5 7B & \cite{liu-etal-2023-LLaVA1.5} & Llama2 & 7B & llava-hf/llava-1.5-7b-hf \\
     LLaVA-1.6 7B & \cite{liu-etal-2024-LLaVA1.6} & Mistral & 7B & llava-hf/llava-v1.6-mistral-7b-hf \\
     LLaVA-CALM2-SigLIP & \cite{inagaki-llava-calm2-siglip} & CALM2 & 7B & cyberagent/llava-calm2-siglip \\
     Japanese Stable VLM & \cite{shing-akiba-jStableVLM} & Japanese Stable LM Instruct Gamma & 7B & stabilityai/japanese-stable-vlm \\ 
     Japanese InstructBLIP Alpha & \cite{shing-akiba-jInstructBLIPAlpha} &  Japanese StableLM Instruct Alpha & 7B & stabilityai/japanese-instructblip-alpha \\
     Llama-3-EvoVLM-JP-v2 & \cite{akiba-etal-2024-evomodelmerge} & Merged \breakInTable{Mantis-8B-SigLIP-Llama-3 \\
        Llama-3-ELYZA-JP-8B \\
        Bunny-v1.1-Llama-3-8B-V}
     & 8B & SakanaAI/Llama-3-EvoVLM-JP-v2 \\ 
    \midrule
    GPT-4o & \multirow{1}{*}{\cite{openai-2024-gpt4}} & \multirow{1}{*}{GPT-4} & - & gpt-4o-2024-05-13\\
    \bottomrule
    \end{tabular}
}
\label{table:model_information}
\end{table*}

%% file: tables/prompt_llm_as_a_judge.tex
\begin{table*}[h]
    \centering
    \footnotesize
    \caption{
      The prompt used in LLM-as-a-judge process. \{input\_text\}, \{answer\}, and \{pred\} indicate the place to insert the question, answer and VLM's prediction, respectively.
    }
    \label{tab:app-prompt-llm-as-a-judge}
    \scalebox{0.96}{
        \begin{tabularx}{\linewidth}{X}
          \toprule
            \textbf{Prompt} \\
          \midrule
            \input{tables/template} \\
          \bottomrule
        \end{tabularx}
    }
    
\end{table*}

%% file: tables/template.tex
\# Instruction

You are an expert evaluator. Your task is to evaluate the quality of the responses generated by AI models.
We will provide you with the user prompt and an AI-generated responses.
You should first read the user prompt carefully for analyzing the task, and then evaluate the quality of the responses based on and rules provided in the Evaluation section below.

\# Evaluation

\#\# Metric Definition

You will be assessing question answering quality, which measures the overall quality of the answer to the question in the user prompt. Pay special attention to length constraints, such as in X words or in Y sentences. The instruction for performing a question-answering task is provided in the user prompt. The response should not contain information that is not present in the context (if it is provided).

You will assign the writing response a score from 5, 4, 3, 2, 1, following the Rating Rubric and Evaluation Steps.
Give step-by-step explanations for your scoring, and only choose scores from 5, 4, 3, 2, 1.

\#\# Criteria Definition

Instruction following: The response demonstrates a clear understanding of the question answering task instructions, satisfying all of the instruction's requirements.

Groundedness: The response contains information included only in the context if the context is present in the user prompt. The response does not reference any outside information.

Completeness: The response completely answers the question with sufficient detail.

Fluent: The response is well-organized and easy to read.

\#\# Rating Rubric

5: (Very good). The answer follows instructions, is grounded, complete, and fluent.

4: (Good). The answer follows instructions, is grounded, complete, but is not very fluent.

3: (Ok). The answer mostly follows instructions, is grounded, answers the question partially and is not very fluent.

2: (Bad). The answer does not follow the instructions very well, is incomplete or not fully grounded.

1: (Very bad). The answer does not follow the instructions, is wrong and not grounded.

\#\# Evaluation Steps

STEP 1: Assess the response in aspects of instruction following, groundedness,completeness, and fluency according to the criteria.

STEP 2: Provide overall score based on the rubric in the format of `Score: X` where X is the score you assign to the response.

\# Question, Reference Answer, and AI-generated Response

\#\# Question

\{input\_text\}

\#\# Reference Answer

\{answer\}

\#\# AI-generated Response

\{pred\}

%% file: tables/table_heron.tex
\begin{table*}[t]
\centering
\small
\begin{tabular}{lcccc}
\toprule
& \multicolumn{4}{c}{\textbf{Japanese Heron-Bench}} \\ 
& \textbf{Detail} & \textbf{Conv} & \textbf{Complex} & \textbf{Average}  \\
\midrule
Japanese InstructBLIP Alpha         & 12.4 & 13.9 & 15.7 & 14.0\\
Japanese Stable VLM                & 18.9 & 30.7 & 23.0 & 24.2\\ 
Llama-3-EvoVLM-JP-v2                  & 43.1 & 37.9 & 36.9 & 39.3\\ 
LLaVA-CALM2-SigLIP                 & 45.4 & 45.8 & 38.8  & 43.3\\
\midrule
LLaVA-1.6 7B              & 21.3 & 27.5 & 28.7 & 25.8\\
LLaVA-1.5 7B                      & 34.7 & 33.8 & 35.7 & 34.8\\
Llama 3.2 11B Vision                 & 34.4 & 40.0 & 35.1 & 36.5\\
InternVL2 8B                        & 48.9 & 41.1 & 45.5 & 45.2\\
Qwen2-VL 7B Inst                     & 57.2 & 54.2 & 53.1 & 54.8\\
\midrule
\textbf{VILA-jp (Ours)} step2 translated   & 46.3 & 43.4 & 52.0 & 47.2\\
\textbf{VILA-jp (Ours)} no step1          & \textbf{60.5} & 50.3 & 58.8 & 56.5\\
\textbf{VILA-jp (Ours)} no interleaved     & 57.7 & \textbf{54.9} & \textbf{63.3} & \textbf{58.6} \\
\textbf{VILA-jp (Ours)} Step 0\&1\&2      & 57.2 & 54.4 & 60.0 & 57.2 \\
\midrule
GPT-4o       & 94.3 & 80.4 & 88.2 & 87.6\\
\bottomrule
\end{tabular}
\caption{Performance of VLMs on Japanese Heron Bench dataset. }
\label{table:heron}
\end{table*}